\documentclass[sigconf]{acmart}
 
\usepackage{bm}
\usepackage{multirow}
\usepackage{paralist}
\copyrightyear{2022}
\acmYear{2022}
\setcopyright{acmcopyright}
\acmConference[MM '22]{Proceedings of the 30th ACM International Conference on Multimedia}{October 10--14, 2022}{Lisbon, Portugal} 
\acmBooktitle{Proceedings of the 30th ACM International Conference on Multimedia (MM '22), October 10--14, 2022, Lisbon, Portugal}

\begin{document}
\fancyhead{}

\title{VMRF: View Matching Neural Radiance Fields}

\author{Jiahui Zhang}
\email{jiahui003@e.ntu.edu.sg}
\affiliation{%
  \institution{Nanyang Technological University}
  \country{Singapore}
}
\author{Fangneng Zhan}
\email{fzhan@mpi-inf.mpg.de}
\affiliation{%
  \institution{Max Planck Institute for Informatics}
  \city{Saarbr\"ucken}
  \state{Saarland}
  \country{Germany}
}
\author{Rongliang Wu}
\email{ronglian001@e.ntu.edu.sg}
\affiliation{%
  \institution{Nanyang Technological University}
  \country{Singapore}
}

\author{Yingchen Yu}
\email{yingchen001@e.ntu.edu.sg}
\affiliation{%
  \institution{Nanyang Technological University \& Alibaba Group}
  \country{Singapore}
}

\author{Wenqing Zhang}
\email{wenqingzhang@bytedance.com}
\affiliation{%
  \institution{ByteDance}
  \country{Singapore}
}

\author{Bai Song}
\email{songbai.site@gmail.com}
\affiliation{%
  \institution{ByteDance}
  \country{Singapore}
}

\author{Xiaoqin Zhang}
\email{zhangxiaoqinnan@gmail.com}
\affiliation{%
  \institution{Wenzhou University}
  \country{China}
}

\author{Shijian Lu}
\authornote{Corresponding author.}
\email{shijian.lu@ntu.edu.sg}
\affiliation{%
  \institution{Nanyang Technological University}
  \country{Singapore}
}

\renewcommand{\shortauthors}{Zhang, et al.}

\begin{abstract}

Neural Radiance Fields (NeRF) has demonstrated very impressive performance in novel view synthesis via implicitly modelling 3D representations from multi-view 2D images. However, most existing studies train NeRF models with either reasonable camera pose initialization or manually-crafted camera pose distributions which are often unavailable or hard to acquire in various real-world data. We design \textit{VMRF}, an innovative view matching NeRF that enables effective NeRF training without requiring prior knowledge in camera poses or camera pose distributions. VMRF introduces a view matching scheme, which exploits unbalanced optimal transport to produce a feature transport plan for mapping a rendered image with randomly initialized camera pose to the corresponding real image. With the feature transport plan as the guidance, a novel pose calibration technique is designed which rectifies the initially randomized camera poses by predicting relative pose transformations between the pair of rendered and real images. Extensive experiments over a number of synthetic and real datasets show that the proposed VMRF outperforms the state-of-the-art qualitatively and quantitatively by large margins. 
\end{abstract}



\begin{teaserfigure}
  \includegraphics[width=0.98\textwidth]{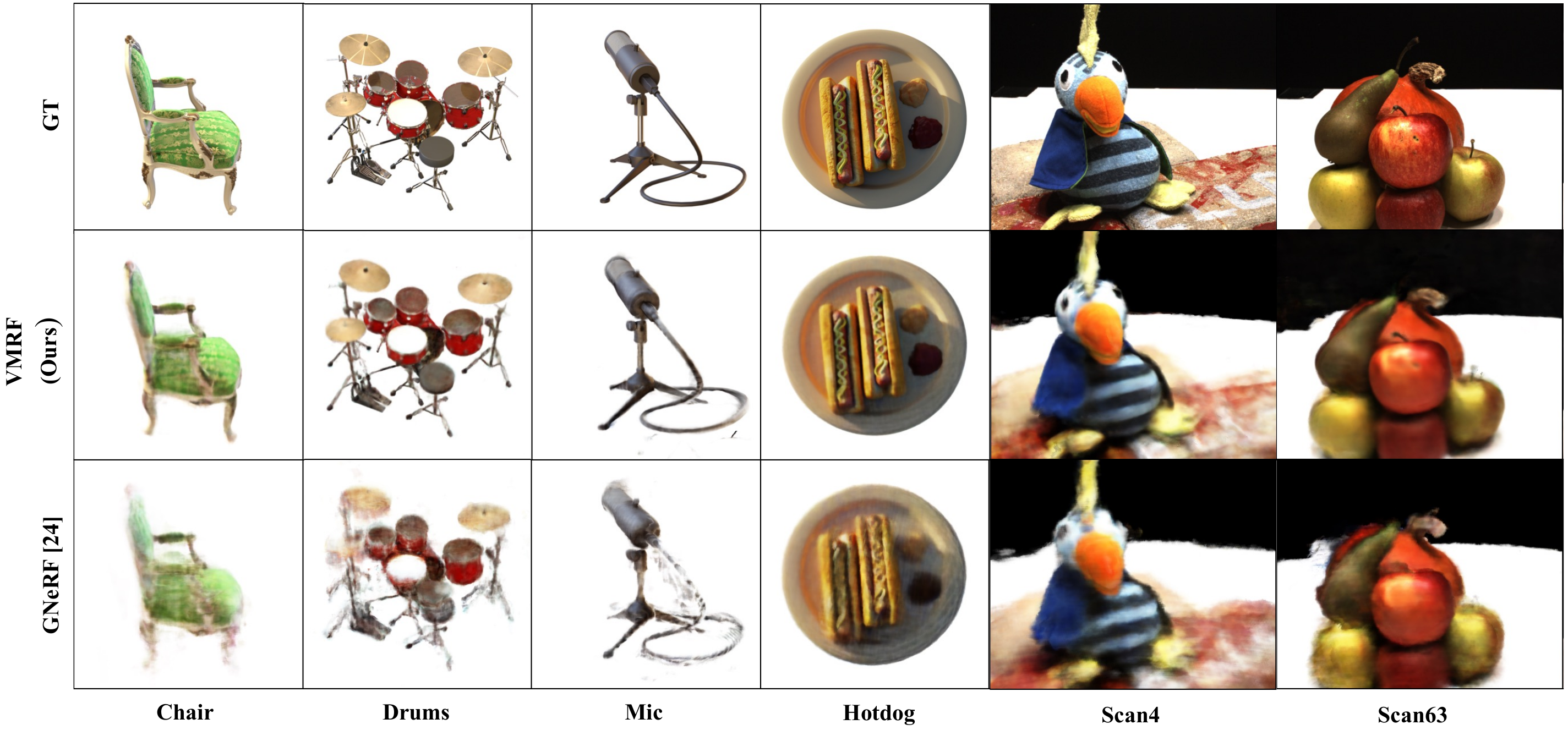}
  \caption{
  Illustration of novel view synthesis by NeRFs (without prior knowledge in camera poses or pose distributions in NeRF training): 
  The state-of-the-art GNeRF~\cite{meng2021gnerf} directly estimates \textit{absolute pose} from image, which is susceptible to out-of-dataset views in training.
  The proposed VMRF instead estimates \textit{relative pose} (with respect to the initially randomized pose) based on view matching results, which is robust to out-of-dataset views and learns better NeRF representations. 
  The samples are from Synthetic-NeRF~\cite{mildenhall2020nerf} and DTU~\cite{jensen2014large}.
  }
  \label{fig:teaser}
\end{teaserfigure}

\maketitle

\section{Introduction}

Neural Radiance Fields (NeRF) \cite{mildenhall2020nerf} has recently been proposed to learn 3D scene representations from multi-view 2D images. By representing a static scene as a continuous 5D function, it can achieve novel view synthesis via volume rendering. However, the NeRF training in most existing methods requires knowledge of accurate camera poses associated with input images, which is often either unavailable or hard to collect in real-world data. Although classical methods such as Structure-from-Motion \cite{hartley2003multiple, schonberger2016structure} can be adopted for pre-computing camera poses, they rely on keypoint detection and cannot recover view-dependent appearance which directly leads to unrealistic novel view rendering especially for scenes with low texture and repetitive patterns \cite{wang2021nerf, meng2021gnerf}.

To reduce the reliance on accurate camera poses, Wang et al. \cite{wang2021nerf} modify the original NeRF \cite{mildenhall2020nerf} for joint optimization of 3D scene representations and camera poses. In addition, Lin et al. \cite{lin2021barf} design bundle-adjusting neural radiance fields for training NeRF from imperfect camera poses. However, both the two methods require camera pose initialization with reasonable accuracy. Meng et al. \cite{meng2021gnerf} propose GNeRF that optimizes NeRF representations with an inversion network for direct pose estimation. It entails an accurate pose distribution with which random camera poses can be sampled for training the inversion network for directly generating poses from input images. However, the accurate pose distribution is often unavailable and needs to be handcrafted for specific datasets. Without this accurate pose distribution, views that are beyond the training dataset would disturb the training of inversion network, which further compromise the optimization of NeRF representations \cite{meng2021gnerf}.

In this paper, we propose an innovative \textbf{V}iew \textbf{M}atching Ne\textbf{RF} (\textbf{VMRF}) that learns NeRF representations without prior knowledge in camera poses or hand-crafted camera pose distributions. 
Instead of estimating camera poses from images directly, we design a novel relative pose estimation strategy that mitigates the dependence on hand-crafted camera pose distributions effectively. Specifically, with random initialization of camera poses, we design a novel view matching scheme that leverages optimal transport \cite{peyre2019computational} to generate a transport plan that represents the feature matching between a real image and the corresponding rendering with the initial camera poses. To tackle non-surjective matching between different views, we introduce unbalanced optimal transport that achieves effective transport in the presence of unequal distribution masses. With feature correspondences from the obtained transport plan, a matching-based pose calibration technique with a relative transformation predictor is designed which learns \textit{relative} pose transformation to calibrate the initially randomized poses towards the corresponding real-image poses. 
Thanks to the \textit{relative} pose estimation strategy, VMRF learns accurate NeRF representations and achieves superior novel view synthesis as illustrated in Fig.~\ref{fig:teaser}.

The contributions of this work are threefold. \textit{First}, we propose VMRF, a novel view matching NeRF that achieves superior NeRF representations without prior knowledge in camera poses or hand-crafted camera pose distributions. 
\textit{Second}, we design a view matching scheme that exploits unbalanced optimal transport to build feature correspondences between cross-view images effectively.
\textit{Third}, we design a matching-based pose calibration technique that yields accurate camera poses by predicting relative pose transformation between real and randomly rendered images.

\section{Related Work}
\paragraph{\textbf{Neural Radiance Fields (NeRF)}} Recently, Mildenhall et al. \cite{mildenhall2020nerf} propose NeRF to achieve novel view synthesis by learning 3D scene representations from a set of posed input images. The key idea is to optimize a 5D scene function (represented by multi-layer perceptron (MLP)), which first maps 3D location and 2D viewing direction to RGB color and volume density and then utilizes differentiable volume rendering to accumulate colors and densities into 2D images from arbitrary viewpoints. Due to the multi-view consistency of the generated images, NeRF as well as its variants \cite{zhang2020nerf++, barron2021mip, liu2020neural, schwarz2020graf, gu2021stylenerf, chen2021mvsnerf, martin2021nerf} has attracted increasing attention in different tasks such as dynamic scene representation \cite{park2021nerfies, du2021neural, gao2021dynamic, guo2021ad, tretschk2021non}, color and shape editing \cite{liu2021editing, yang2021learning}, compositional scene modeling \cite{niemeyer2021giraffe}, scene relighting \cite{srinivasan2021nerv, boss2021nerd} and skeleton-driven synthesis \cite{peng2021animatable}. However, most existing NeRF-related designs require accurate camera poses for reliable NeRF training.

Several attempts have been made for relaxing the requirement for accurate camera poses. For example, Wang et al. \cite{wang2021nerf} firstly propose to optimize camera poses and NeRF representations jointly. Lin et al. \cite{lin2021barf} design a coarse-to-fine bundle adjustment technique to learn NeRF representations from imperfect camera poses. Jeong et al. \cite{jeong2021self} propose a camera self-calibration algorithm for fine-tuning the initialized camera poses for improving NeRF representations. However, these methods tend to suffer severe performance degradation while working with rough pose initialization. Meng et al. \cite{meng2021gnerf} propose to learn NeRF representations with random camera pose initialization, where an inversion network is introduced to directly
estimate camera poses from images. However, it relies on accurate camera pose distributions which are dataset-specific and hand-crafted. Differently, the proposed VMRF allows to optimize NeRF representations with no knowledge of camera poses or hand-crafted camera pose distributions, thanks to our designed view matching scheme and matching-based camera pose calibration technique.

\paragraph{\textbf{Camera Pose Estimation for NeRF}} As reliable NeRF training requires accurate camera poses that are associated with the input images, most existing NeRF-based methods exploit classical Structured from Motion (SfM) techniques \cite{hartley2003multiple, wu2013towards, schonberger2016structure}) to build up and estimate per-scene camera poses based on keypoints detection and matching. However, SfM tends to fail in scenes with low texture and repetitive patterns, where few keypoints can be detected. With the recent advance of pose-free NeRF (i.e., NeRF trained without accurate camera poses), learning-based camera pose estimation \cite{lin2021barf, meng2021gnerf} has attracted increasing attention for estimating camera poses of input real images for NeRF training. A representative work is inversion network in GNeRF \cite{meng2021gnerf}, which is trained to invert the input image to the corresponding camera pose. However, the training of the inversion network is susceptible to new views (beyond the training dataset) when hand-crafted camera pose distributions of reasonable accuracy are unavailable for camera pose initialization \cite{meng2021gnerf}. Instead of predicting camera poses directly from images, VMRF exploits view matching results to learn relative pose transformation which calibrates randomly initialized camera poses towards accurate camera pose with no prerequisite for either camera poses or hand-crafted camera pose distributions.

\paragraph{\textbf{Optimal Transport}} Optimal transport (OT) \cite{villani2009optimal} is proposed to compare different distributions and yield an optimal transport plan for distribution matching. In recent years, optimal transport has been widely explored in the computer vision community \cite{courty2014domain,kolkin2019style,zhan2022gmlight,zhan2021sparse,zhan2021emlight}. For example, Liu et al. \cite{liu2020semantic} exploit optimal transport in a unified framework to model semantic correspondences. For tackling unbalanced distributions with different masses and deviations, unbalanced optimal transport \cite{chizat2018scaling, liero2018optimal} has also been explored for different tasks, such as image translation \cite{zhan2021unbalanced, zhan2022modulated, zhan2021bi}. To the best of our knowledge, the proposed VMRF the first that adapt unbalanced optimal transport for optimizing NeRF representations.

\begin{figure*}[t]
\centering
\includegraphics[width=0.936\linewidth]{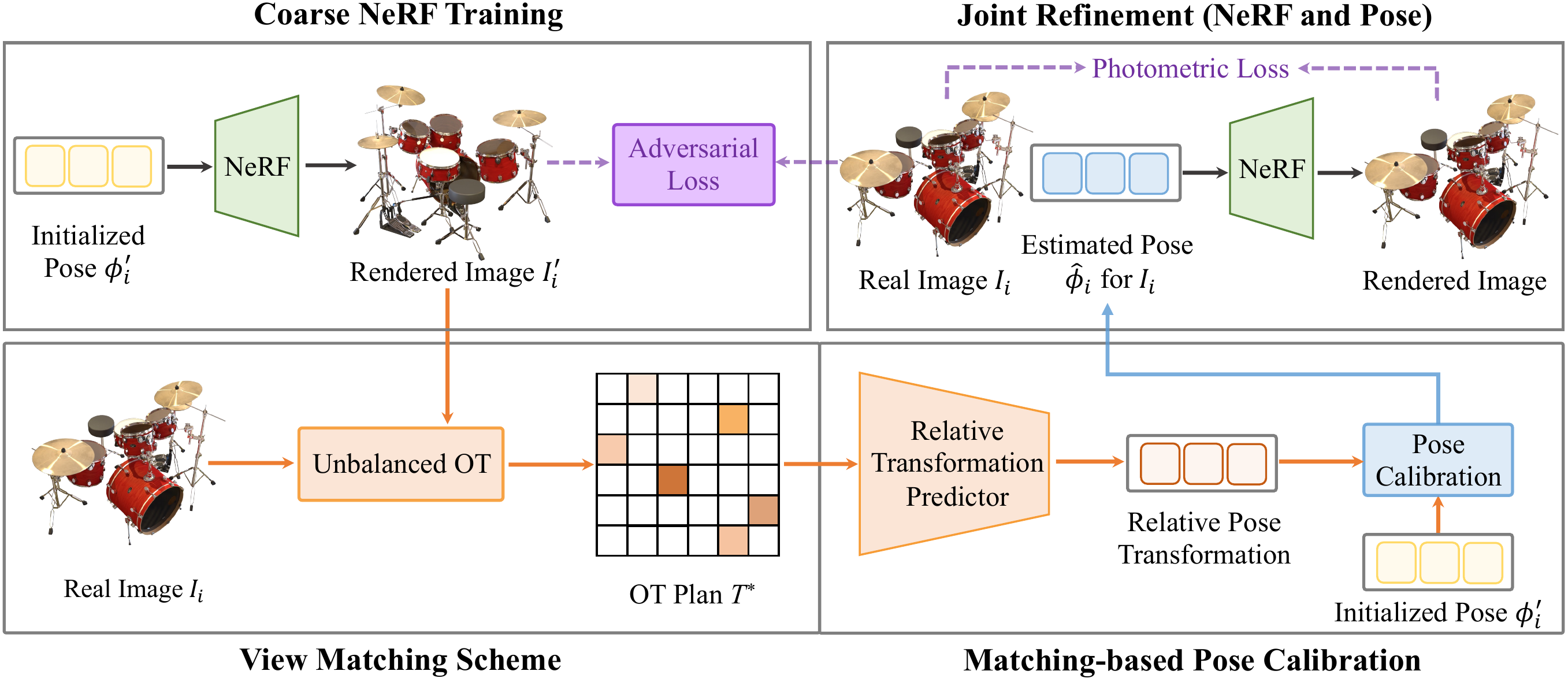}
\caption{
The framework of VMRF: A coarse NeRF is first trained with randomly \textit{Initialized Poses} as in \cite{meng2021gnerf}. Camera poses of real images are then estimated via the proposed view matching scheme and matching-based pose calibration. Specifically, view matching learns to build an \textit{Optimal Transport (OT) Plan} $T^*$ between the image $I'_i$ rendered by the initial pose $\phi'_i$ and the corresponding real image $I_i$. The relative pose transformation between the two images is then estimated from $T^*$, with which the pose $\hat{\phi}_i$ of $I_i$ can be computed by calibrating the initial camera pose $\phi'_i$. Thus, camera pose $\hat{\phi}_i$ and real image $I_i$ can be paired for NeRF training. Meanwhile, $\hat{\phi}_i$ and the coarse NeRF are jointly refined with a photometric loss. Similar to GNeRF \cite{meng2021gnerf}, VMRF is end-to-end trainable under a hybrid and iterative optimization scheme.
}
\label{vmrf}
\end{figure*}

\section{Preliminary}

This section presents the preliminary knowledge in camera pose distributions and NeRF 3D representations.

\subsection{Camera Pose Distribution} 
\label{Pose Distribution}
Camera pose distribution imposes a range on random initialization of camera poses, which can be determined by five parameters \cite{meng2021gnerf}: the radius of the sphere where cameras are distributed, the range of camera elevation, the range of camera azimuth, camera lookat points and camera up vector.
Well aligned with the finding in GNeRF \cite{meng2021gnerf}, we observe that novel view synthesis is largely affected by the variation in the range of camera azimuth and elevation (more details provided in the supplementary material). We therefore focus on studying the range of camera azimuth and elevation (via random selection with no handcrafted priors) in camera pose distribution in this work.

\subsection{NeRF 3D Representation} 

We adopt the NeRF in \cite{mildenhall2020nerf} to represent 3D scenes. A neural radiance field is a continuous 5D function that maps an input 3D point $x \in \mathbb{R}^3$ and a 2D viewing direction $d \in \mathbb{S}^2$ to an RGB color $c \in \mathbb{R}^3$ and volume density $\sigma \in \mathbb{R}$. It can be formulated as $F_\Theta(x,b) \rightarrow (c, \sigma)$, where $F$ is parameterized with MLP and $\Theta$ are network parameters. To render a 2D image from 3D NeRF representations, a differentiable volume renderer is adopted with a numerical integrator \cite{mildenhall2020nerf} for approximating the intractable volumetric projection integral. By taking $N$ random samples along a camera ray $r$ and representing their color and volume density values as $\{(c_r^i, \sigma_r^i)\}_{i=1}^N$, the RGB color $c_r$ in 2D image can be rendered by:
\begin{equation}
    c_r = \sum_{i=1}^N T_r^i\alpha_r^i c_r^i \qquad
    T_r^i = \prod_{j=1}^{i-1}(1 - \alpha_r^j) \qquad
    \alpha_r^i = 1 - exp(\sigma_r^i \delta_r^i)
\end{equation}

where $T_r^i$ and $\alpha_r^i$ represent the transmittance and alpha value of $i$-th sample along camera ray $r$, respectively, and $\delta_r^i = x_r^{i+1} - x_r^i$ is the distance between adjacent samples. Given a dataset of RGB images $\mathcal{I} = \{I_1, I_2, ... ,I_n\}$ of a static scene with their corresponding camera poses $\Phi = \{\phi_1, \phi_2, ..., \phi_n \}$, NeRF optimizes parameters $\Theta$ by a photometric loss (in the form of the sum of squared differences) between the real images $\mathcal{I}$ and the corresponding rendered images. 

\section{Proposed Method}

\subsection{Pipeline}

VMRF learns effective NeRF representations without prior knowledge in camera poses or camera pose distributions. Similar to GNeRF \cite{meng2021gnerf}, it first learns a coarse NeRF with randomly initialized camera poses $\Phi' = \{\phi'_i, i \in [1, n]\}$ as illustrated in Fig.~\ref{vmrf}. With the coarse NeRF, camera poses $\hat{\Phi}=\{\hat{\phi}_i, i \in [1, n]\}$ of real images $\mathcal{I} = \{I_i, i \in [1, n]\}$ are then estimated via view matching and matching-based pose calibration. Specifically, the view matching yields an optimal transport plan by matching the views of a real image and the corresponding rendering with the randomly initialized camera pose. The relative pose transformation between the real and the rendered images are then derived from the optimal transport plan by the \textit{Relative Transformation Predictor}. The camera pose of the real image (i.e., $\hat{\phi}_i$) can thus be determined for NeRF training. Note that we also employ a photometric loss to jointly refine the NeRF training as well as the estimated camera poses as illustrated in \textit{Joint Refinement}. VMRF utilizes a hybrid and iterative optimization scheme as in GNeRF \cite{meng2021gnerf} to achieve end-to-end training.

 
\subsection{View Matching Scheme}
\label{View Matching}

The view matching scheme aims to match features between rendered images with randomly initialized camera poses and real images, thus providing accurate guidance for the subsequent camera pose calibration. To avoid many-to-one matching and ensure end-to-end optimization, we introduce optimal transport with entropic regularization that can achieve image feature matching in a differentiable way.

Standard optimal transport aims at computing a minimal cost transportation (i.e., optimal transport plan) to transform a source distribution into a target distribution with the same total mass. In our work, we model feature matching between images from different views as an optimal transport problem. The image rendered with randomly initialized pose and the real image are thus treated as the source image and the target image, respectively. Specifically, we first adopt two pre-trained VGG-19 \cite{simonyan2014very} to extract source feature set $\boldsymbol{f^s} \in \mathbb{R}^{l \times d}$ and target feature set $\boldsymbol{f^t} \in \mathbb{R}^{l \times d}$ from the source and target images, respectively, where each feature set involves $l$ feature vectors and $d$ denotes the dimension of feature vectors. Then, we encode the two feature sets to build discrete source and target distributions ($\mu_s$ and $\mu_t$) as follows:
\begin{equation}
    \mu_s = \sum_{i=1}^{l} p_i^s\delta(f^s_i), \qquad \qquad \mu_t = \sum_{j=1}^{l} p_j^t\delta(f^t_j),
\end{equation}
where $\delta(\cdot)$ denotes the Dirac function, and $p_i^s$ and $p_j^t$ represent the probability mass of the $i$-th source feature vector $f_i^s$ and the probability mass of the $j$-th target feature vector $f_j^t$, respectively. Then, we define a cost matrix $\boldsymbol{M} \in \mathbb{R}^{l \times l}$ based on matching similarities between the source and target feature vectors, where each element $M_{ij}$ denotes the moving cost from probability mass $p_i^s$ to $p_j^t$. The cost matrix can be computed by: $\boldsymbol{M} = 1 - \frac{\boldsymbol{f^s} \cdot {\boldsymbol{f^t}}^\mathrm{T}}{\left\|\boldsymbol{f^s}\right\| \left\|\boldsymbol{f^t}\right\|}$. The optimal transport problem can thus be formulated by:
\begin{equation}
\label{ot}
\begin{aligned}
    & T^* = \mathop{\arg\min}_{T} \sum_{i,j=1}^{l} T_{ij} M_{ij} \\
    & s.t. \quad T\boldsymbol{1_{l}} = \mu_s, \quad T^\mathrm{T}\boldsymbol{1_{l}} = \mu_t
\end{aligned}
\end{equation}
where $T^*$ is defined as the optimal transport plan and $T_{ij}$ represents the optimal amount of mass moved from $f_i^s$ to $f_j^t$ in order to minimize the total moving cost. Many-to-one matching is effectively circumvented by constraining $\mu_s$ to be the row sum of $T$ and constraining $\mu_t$ to be the column sum of $T$. 

A key constraint of the standard optimal transport is that the total masses of the source and target distributions need to be equal. Due to the occlusion of certain features caused by camera poses, feature matching between images from different views is not surjective and so the source and target distributions have inconsistent total mass. Directly applying the standard optimal transport will thus match all features including outlier samples (i.e., features that exist only in one view), which leads to inaccurate matching and false matching. We introduce unbalanced optimal transport (UOT) \cite{chizat2018scaling} for accurate matching distributions with arbitrary masses. Specifically, we replace the `strict' constraints on the marginals of plan $T$ (in Eq.\ref{ot}) by `soft' penalties with Kullback-Leibler divergence. The unbalanced optimal transport can be formulated as follows:
\begin{equation}
    T^* = \mathop{\arg\min}_{T} \Big[ \sum_{i,j=1}^{l} T_{ij} M_{ij} + \epsilon KL(T\boldsymbol{1_{l}}||\mu_s) + \epsilon KL(T^\mathrm{T}\boldsymbol{1_{l}}||\mu_t) \Big]
\end{equation}
where $\epsilon \textgreater 0$ denotes the regularization parameter.

To implement a differentiable UOT, we introduce entropic regularization for approximating the UOT as follows: 
\begin{equation}
\begin{aligned}
    T^* = 
    &\mathop{\arg\min}_{T} \Big[ \sum_{i,j=1}^{l} T_{ij} M_{ij} - \eta H(T) \\
    & + \epsilon KL(T\boldsymbol{1_{l}}||\mu_s) + \epsilon KL(T^\mathrm{T}\boldsymbol{1_{l}}||\mu_t) \Big],
\end{aligned}
\end{equation}
where $H(T)$ is an entropic regularization term which is defined by: $H(T) = -\sum_{i,j=1}^{l} T_{ij}log(T_{ij} - 1)$, and $\eta \textgreater 0$ is a regularization parameter. 

To obtain optimal transport plan of the entropic regularized UOT, we represent the entropic regularized UOT in the Fenchel-Legendre dual form and introduce the Sinkhorn algorithm \cite{chizat2018scaling} to compute the optimal solution ($u^*, v^*$).
\begin{equation}
\begin{aligned}
    B(u^*, v^*) = T^* = & \mathop{\arg\min}_{u,v \in \mathbb{R}^{l}} \Big[
    \eta  \sum_{i,j=1}^{l} exp(\frac{u^i + v^j - M_{ij}}{\eta}) \\
    & + \epsilon \langle e^{-u/\epsilon}, \mu_s \rangle + \epsilon \langle e^{-v/\epsilon}, \mu_t \rangle \Big],
\end{aligned}
\end{equation}
where $u, v \in \mathbb{R}^l$ denote dual vectors. The optimal transport plan $T^*$ in the dual form can thus be represented by:
\begin{equation}
B(u^*, v^*) = diag(e^{u^*/\eta})e^{M/\eta}diag(e^{v^*/\eta})
\end{equation}

\subsection{Matching-based Pose Calibration}
\label{Matching-based Pose Calibration}

With optimal transport plans between real images and images rendered with randomly initialized poses, we design a matching-based pose calibration technique that estimates the pose of the real image by predicting a relative pose transformation. It involves two major steps as illustrated in Fig~\ref{vmrf}.

First, with feature-level matching (from the obtained transport plan) between a real image and a rendered image, we introduce a relative transformation predictor that predicts relative rotation and relative translation for pose calibration. This process can be formulated as follows:
\begin{equation}
[t_x, t_y, t_z, \theta_x, \theta_y, \theta_z] = RTP(T^*),
\end{equation}
where $RTP(\cdot)$ denotes the relative transformation predictor, $t_x, t_y, t_z$ represent relative translations along the $x$, $y$, and $z$ axes, respectively, and $\theta_x, \theta_y, \theta_z$ represent relative rotations about the $x$, $y$ and $z$ axes, respectively.

We then estimate the camera pose $\hat{\phi}_i$ of the real image by calibrating the randomly initialized camera pose $\phi'_i$ of the rendered image with the predicted relative rotation and relative translation. Specifically, we construct a relative rotation matrix $\Delta \boldsymbol{R}$ and a translation vector $\Delta \boldsymbol{t}$ as follows:

\setlength{\arraycolsep}{1.3pt}
\begin{gather}
\Delta \boldsymbol{R} = {
\left[ \begin{array}{ccc}
1 & 0 & 0\\
0 & cos\theta_x & sin\theta_x \\
0 & -sin\theta_x & cos\theta_x \\
\end{array} 
\right ]}
{
\left[ \begin{array}{ccc}
cos\theta_y & 0 & -sin\theta_y\\
0 & 1 & 0\\
sin\theta_y & 0 & cos\theta_y\\
\end{array} 
\right ]}
{
\left[ \begin{array}{ccc}
cos\theta_z & sin\theta_z & 0 \\
-sin\theta_z & cos\theta_z & 0 \\
0 & 0 & 1  \\
\end{array} 
\right ]}\\
\Delta \boldsymbol{t} = [t_x, t_y, t_z]
\end{gather}
Given the randomly initialized pose $\phi'_i$, the estimated camera pose $\hat{\phi}_i$ associated with the real image can be obtained by:
\begin{equation}
\hat{\phi}_i = {
\left[ \begin{array}{cc}
\Delta \boldsymbol{R} & \Delta \boldsymbol{t}^\mathrm{T}\\
0^3 & 1 \\
\end{array} 
\right ]} \phi'_i
\label{calibration}
\end{equation}

\subsection{Loss Function}
\label{loss function}

In this section, we detail loss functions that are employed in the training process as below.

\paragraph{\textbf{Adversarial loss}} 
Due to the lack of known camera poses, we follow \cite{meng2021gnerf} and first train a coarse NeRF with randomly initialized poses and an adversarial loss. The images generated by the coarse NeRF have similar distribution as that of real images in training dataset. The adversarial loss $\mathcal{L}_{adv}$ can be defined as follows:
\begin{equation}
\begin{aligned}
    \mathcal{L}_{adv}(F, D) & =  \mathbb{E}_{I\sim P_{data}}[log(D(I))]\\
    & + \mathbb{E}_{F(\Phi')\sim P_{g}}[log(1 - D(F(\Phi')))], 
\end{aligned}
\end{equation}
where $F$ and $D$ represent the NeRF model and the discriminator, $P_{data}$ and $P_g$ denote the distribution of real images and generated images, and $\Phi'$ represents a set of randomly initialized poses.

\paragraph{\textbf{Calibration loss}} We introduce a calibration loss $\mathcal{L}_{ca}$ to train the relative transformation predictor to estimate relative pose transformation from the obtained transport plan. This is performed in an unsupervised manner as follows:
\begin{equation}
\begin{aligned}
    & \mathcal{L}_{ca}(RTP) = \mathbb{E}\Big[\left\| F_{ca}(RTP(T^{*}_{ab}), \phi'_{a}) - \phi'_{b} \right\|_{2}^2 \Big] \\
    &= \mathbb{E}\Big[\left\| F_{ca}(RTP(F_{uot}(F(\phi'_{a}),F(\phi'_{b}))), \phi'_{a}) - \phi'_{b} \right\|_{2}^2 \Big] 
\end{aligned}
\end{equation}
where ($\phi'_{a}$, $\phi'_{b}$) is a pair of poses with random initialization, $T^{*}_{ab}$ denotes optimal transport plan between images rendered with $\phi'_{a}$ and $\phi'_{b}$, $F_{ca}(\cdot)$ represents the calibration process described in Eq.~\ref{calibration}, $F_{uot}$ denotes the unbalanced optimal transport, and $RTP(\cdot)$ denotes our relative transformation predictor. 


\paragraph{\textbf{Photometric loss}} For the joint refinement of NeRF model $F$ and the estimated camera poses $\hat{\Phi}$, a photometric loss $\mathcal{L}_P$ is employed which is formulated as follows:
\begin{equation}
\mathcal{L}_P(F, \hat{\Phi} ) = \frac{1}{n} \sum_{i=1}^{n} \left\| I_i - F(\hat{\phi}_i) \right\|_{2}^2
\end{equation}
Note, the VMRF is trained end-to-end following the hybrid and iterative optimization scheme in GNeRF \cite{meng2021gnerf}.

\subsection{Network Details}
\label{implementation details}

In the view matching scheme, The number of feature vector in each feature set is set to $l=64$ and the dimension of feature vector is set to $d=256$. The regularization parameter $\eta$ that controls the smoothness of UOT is set to $0.005$, parameter study of $\eta$ is presented in Section~\ref{Para_investi}. For the matching-based pose calibration, we utilize the vision transformer \cite{dosovitskiy2020image} as relative transformation predictor, where the last layer is modified to generate relative transformation values. For camera pose, we employ a 3D embedding in Euclidean space as camera position. Following \cite{zhou2019continuity}, we represent camera rotation as a continuous 6D embedding. In term of NeRF in VMRF, we utilize the original NeRF architecture \cite{mildenhall2020nerf}. 

\section{Experiment}

\subsection{Datasets and Implementation Details}
\label{dataset}

\begin{table}[t]
\small 
    \renewcommand\tabcolsep{3.5pt}
	\centering
	\caption{Quantitative comparisons of novel view synthesis on the dataset Synthetic-NeRF: VMRF outperforms the state-of-the-art GNeRF \cite{meng2021gnerf} consistently in PSNR, SSIM and LPIPS for different synthetic scenes with no hand-crafted pose distributions. $D_{s1}$, $D_{s2}$ and $D_{s3}$ denote randomly selected pose distributions, where the ranges of azimuth and elevation ($R_{azi}, R_{ele}$) are ($[0^\circ, 360^\circ]$, $[-30^\circ, 90^\circ]$), ($[0^\circ, 360^\circ]$, $[-60^\circ, 90^\circ]$) and ($[0^\circ, 360^\circ]$, $[-90^\circ, 90^\circ]$), respectively. GNeRF$^*$ denote a GNeRF variant that has the same NeRF settings and similar number of parameters as the proposed VMRF. All methods are trained with the same training data and batch size. 
	}
	\label{tab_synthetic}
	\begin{tabular}{*{8}{c}}
		\toprule
		\multirow{2}{*}{} & \multirow{2}{*}{Scene} & \multicolumn{2}{c}{PSNR$\uparrow$ }  & \multicolumn{2}{c}{SSIM$\uparrow$ } & \multicolumn{2}{c}{LPIPS$\downarrow$ } \\
		
		\cmidrule(lr){3-4}\cmidrule(lr){5-6}\cmidrule(lr){7-8} 
		
		&& \multicolumn{1}{c}{GNeRF$^*$} & \multicolumn{1}{c}{\textbf{VMRF}} &
		\multicolumn{1}{c}{GNeRF$^*$} & \multicolumn{1}{c}{\textbf{VMRF}} &
		\multicolumn{1}{c}{GNeRF$^*$} & \multicolumn{1}{c}{\textbf{VMRF}} \\
		
		\hline
		\multirow{6}{*}{$D_{s1}$} & Chair & 25.01 & \textbf{26.05} & 0.8940 & \textbf{0.9083} & 0.1529 & \textbf{0.1397} \\
		& Drums & 20.63 & \textbf{23.07} & 0.8628 & \textbf{0.8917} & 0.2019 & \textbf{0.1605} \\
		& Lego & 22.95 & \textbf{25.23} & 0.8493 & \textbf{0.8865} & 0.1630 & \textbf{0.1215} \\
		& Mic & 23.68 & \textbf{27.63} & 0.9332 & \textbf{0.9483} & 0.1095 & \textbf{0.0803}\\
		& Ship & 17.91 & \textbf{21.39} & 0.7626 & \textbf{0.7998} & 0.3628 & \textbf{0.2933} \\
		\hline
		\hline
		\multirow{6}{*}{$D_{s2}$} & Chair & 22.18 & \textbf{24.53} & 0.8753 & \textbf{0.8987} & 0.2006 & \textbf{0.1695}  \\
		& Drums & 19.05 & \textbf{21.25} & 0.8312 & \textbf{0.8652} & 0.2670 & \textbf{0.2098}\\
		& Lego & 21.39 & \textbf{23.51} & 0.8412 & \textbf{0.8618} & 0.1794 & \textbf{0.1429} \\
		& Mic & 23.22 & \textbf{24.39} & 0.9216 & \textbf{0.9392} & 0.1301 & \textbf{0.1021}\\
		& Ship & 15.89 & \textbf{20.35} & 0.7416 & \textbf{0.7895} & 0.4034 & \textbf{0.3502} \\
		\hline
		\hline
		\multirow{6}{*}{$D_{s3}$} & Chair & 21.27 & \textbf{23.18} & 0.8665 & \textbf{0.8886} & 0.2326 & \textbf{0.1593} \\
		& Drums & 18.08 & \textbf{20.01} & 0.8132 & \textbf{0.8371} & 0.3314 & \textbf{0.2877} \\
		& Lego & 18.22 & \textbf{21.59} & 0.8176 & \textbf{0.8320} & 0.2363 & \textbf{0.1762} \\
		& Mic & 17.22 & \textbf{20.29} & 0.8592 & \textbf{0.8971} & 0.3154 & \textbf{0.2208} \\
		& Ship & 14.31 & \textbf{17.58} & 0.7295 & \textbf{0.7602} & 0.4673 & \textbf{0.3913}\\
		\bottomrule
	\end{tabular}

\end{table}

\paragraph{\textbf{Datasets }} 

For training and testing, we follow the dataset setting of GNeRF\cite{meng2021gnerf} and conduct experiments on both synthetic and real scenes. For synthetic scenes, we employ the dataset Synthetic-NeRF \cite{mildenhall2020nerf} which consists of objects with complicated geometry. We adopt the original split of the dataset where 100 images are used for training and eight images are randomly sampled from the test set for evaluations. The input image is resized to $400\times 400$ for both training and test. For real scenes, we select three representatives in the dataset DTU~\cite{jensen2014large} including `scan4' and `scan63' that are characterized by rich textures as well as `scan48' that has very sparse texture. For a total of 49 images per scene, we perform the training on 43 images and use the rest 6 images for evaluation. For both training and test, the image size is fixed at $500\times 400$.

\paragraph{\textbf{Implementation Details}} 
For various parameters in camera pose distributions (detailed in Section \ref{Pose Distribution}), we focus on the range of camera elevation and azimuth (via random selection with no handcrafted priors) as the two parameters are decisive factors that affect novel view synthesis (which has been verified in \cite{meng2021gnerf} as well as our studies as detailed in the supplementary material). For other insignificant parameters, we follow GNeRF \cite{meng2021gnerf} and set the sphere radius at $4.0$ for both synthetic and real datasets, and set the camera lookat points at $(0,0,0)$ and $\mathcal{N}(0, 0.01^2)$, respectively, for the synthetic dataset and real dataset. We adopt Adam optimizer to train the NeRF, discriminator, relative transformation predictor and camera poses with initial learning rates of 0.0005, 0.0001, 0.0001 and 0.005, respectively. The mini-batch size is set at 12 for both synthetic and real scenes in training. We use the Pytorch framework in implementation and employ one NVIDIA RTX 3090ti GPU for both training and inference. 

\begin{table}[t]
\small 
    \renewcommand\tabcolsep{3.5pt}
	\centering
	\caption{Quantitative comparisons of novel view synthesis on the dataset DTU: VMRF outperforms the state-of-the-art GNeRF$^*$ \cite{meng2021gnerf} with no hand-crafted pose distributions. $D_{r1}$ and $D_{r2}$ denote randomly selected pose distributions, where the range of azimuth and elevation ($R_{azi}, R_{ele}$) are ($[0^\circ, 180^\circ]$, $[0^\circ, 90^\circ]$) and ($[0^\circ, 180^\circ]$, $[-30^\circ, 90^\circ]$), respectively. All methods are trained with the same training data and batch size. 
	}
	\label{tab_real}
	\begin{tabular}{*{8}{c}}
		\toprule
		\multirow{2}{*}{} & \multirow{2}{*}{Scene} & \multicolumn{2}{c}{$\uparrow$ PSNR}  & \multicolumn{2}{c}{$\uparrow$ SSIM} & \multicolumn{2}{c}{$\downarrow$ LPIPS} \\
		
		\cmidrule(lr){3-4}\cmidrule(lr){5-6}\cmidrule(lr){7-8} 
		
		&& \multicolumn{1}{c}{GNeRF$^*$} & \multicolumn{1}{c}{\textbf{VMRF}} &
		\multicolumn{1}{c}{GNeRF$^*$} & \multicolumn{1}{c}{\textbf{VMRF}} &
		\multicolumn{1}{c}{GNeRF$^*$} & \multicolumn{1}{c}{\textbf{VMRF}} \\
		
		\hline
		
		\multirow{3}{*}{$D_{r1}$}   

		& Scan4 & 17.04 & \textbf{19.51} & 0.6124 & \textbf{0.6503} & 0.4676 & \textbf{0.4205} \\
		& Scan48 & 18.24 & \textbf{21.30} & 0.7743 & \textbf{0.7986} & 0.3459 & \textbf{0.3063} \\
		& Scan63 & 19.43 & \textbf{22.52} & 0.8217 & \textbf{0.8636} & 0.3041 & \textbf{0.2551}\\
		\hline
		\hline
		
		\multirow{3}{*}{$D_{r2}$}   

		& Scan4 & 16.22 & \textbf{18.38} & 0.6079 & \textbf{0.6297} & 0.4955 & \textbf{0.4707} \\
		& Scan48 & 15.81 & \textbf{20.61} & 0.7520 & \textbf{0.7913} & 0.4207 & \textbf{0.3185} \\
		& Scan63 & 17.30 & \textbf{20.59} & 0.8152 & \textbf{0.8538} & 0.3202 & \textbf{0.2839} \\
		

       \bottomrule
	\end{tabular}

\end{table}

\begin{table}[t]
\renewcommand\tabcolsep{9pt}
	\centering
	\caption{Quantitative comparisons of camera pose estimation for NeRF training without hand-crafted pose distributions (on Synthetic-NeRF): Rot and Trans denote mean camera rotation differences and mean camera translation differences, respectively. }
	\label{tab_pose}
	\begin{tabular}{*{5}{c}}
		\toprule
		\multirow{2}{*}{Scene} & \multicolumn{2}{c}{GNeRF$^*$ \cite{meng2021gnerf}} & \multicolumn{2}{c}{Ours} \\
		
		\cmidrule(lr){2-3}\cmidrule(lr){4-5}
		
		& \multicolumn{1}{c}{Rot($^{\circ}$)$\downarrow$} & \multicolumn{1}{c}{Trans$\downarrow$ } & \multicolumn{1}{c}{Rot($^{\circ}$)$\downarrow$ } & \multicolumn{1}{c}{Trans$\downarrow$ } \\
		
		\hline
		Chair & 6.078 & 0.339 & \textbf{4.853} & \textbf{0.283} \\
		Drums & 2.769 & 0.137 & \textbf{1.284} & \textbf{0.081}\\
		Lego & 3.315 & 0.251 & \textbf{2.160} & \textbf{0.164} \\
	    Mic & 3.021 & 0.161 & \textbf{1.394} & \textbf{0.069} \\
		Ship & 31.56 & 1.379 & \textbf{16.89} & \textbf{0.714} \\
		\bottomrule
	\end{tabular}
\end{table}

\begin{figure*}[t]
\begin{center}
\includegraphics[width=0.99\linewidth]{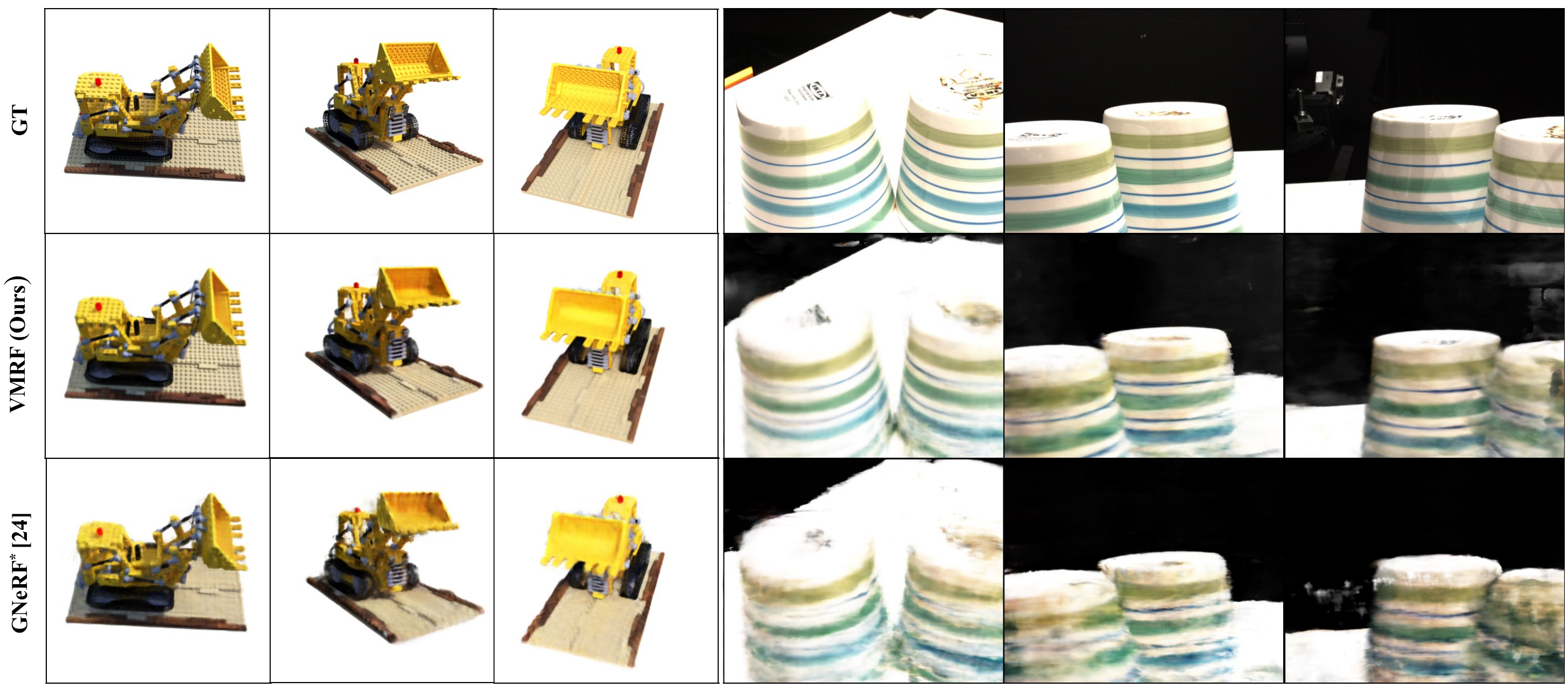}
\end{center}
\caption{
Qualitative comparisons of VMRF with GNeRF$^*$ in novel view synthesis: The comparisons are conducted over different views of the scene 'lego' in Synthetic-NeRF and the scene 'scan48' in DTU, where `GT' denotes the ground-truth images. It is clear that VMRF synthesizes high-fidelity images with less artifacts and finer details compared with GNeRF$^*$.
}
\label{visual_compare}
\end{figure*}

\subsection{Comparisons with the State-of-the-Art}

\paragraph{\textbf{Novel View Synthesis}} We compare VMRF with the state-of-the-art GNeRF \cite{meng2021gnerf} over different synthetic scenes and real scenes. We did not compare with NeRF$--$ \cite{wang2021nerf}, BARF \cite{lin2021barf} and SCNeRF \cite{jeong2021self} as the three methods require reasonable (instead of random) camera pose initialization. In addition, we employ randomly selected pose distributions to verify the effectiveness of the proposed VMRF. Specifically, we set the range of azimuth and elevation ($R_{azi}, R_{ele}$) at ($[0^\circ, 360^\circ]$, $[-30^\circ, 90^\circ]$) to ($[0^\circ, 360^\circ]$, $[-90^\circ, 90^\circ]$) for the dataset Synthetic-NeRF with synthetic scenes. For the dataset DTU with real scenes, ($R_{azi}, R_{ele}$) is set at ($[0^\circ, 180^\circ]$, $[0^\circ, 90^\circ]$) to ($[0^\circ, 180^\circ]$, $[-30^\circ, 90^\circ]$) instead. For fair comparisons, we re-train GNeRF$^*$ \footnote{$*$ indicates a version of GNeRF that has the same NeRF settings and similar number of parameters with VMRF. Details are provided in the supplementary material.} by using its official codes, and both GNeRF$^*$ and VMRF are trained with the same training dataset and the same batch size.

Table \ref{tab_synthetic} and Table \ref{tab_real} show experimental results over datasets Synthetic-NeRF and DTU, respectively, where all evaluations were conducted on the same test images as described in section~\ref{dataset}. We can observe that VMRF outperforms GNeRF$^*$ consistently across PSNR, SSIM and LPIPS as well as all evaluated scenes. The superior performance of VMRF is largely attributed to our designed view matching and pose calibration techniques that conduct effective NeRF training with relative pose estimation without requiring hand-crafted pose distributions. The relative pose estimation is more tolerant to the disturbance resulting from out-of-dataset views which leads to robust training and effective NeRF representations. The quantitative experiments are well aligned with the qualitative experiments as illustrated in Fig.~\ref{visual_compare} where VMRF synthesizes high-fidelity images with less artifacts and finer details.

\paragraph{\textbf{Camera Poses Estimation}} 
We also compare the accuracy of the estimated camera poses used for NeRF training. The comparison is under the setup of random pose distribution selection where the range of azimuth and elevation are randomly set to $[0^\circ, 360^\circ]$ and $[-30^\circ, 90^\circ]$, respectively. Following \cite{meng2021gnerf}, we perform this evaluation on Synthetic-NeRF with the metric of mean camera rotation difference (Rot) and mean translation difference (Trans) that are computed with the toolbox \cite{zhang2018tutorial} on the training set. As Table \ref{tab_pose} shows, VMRF outperforms GNeRF$^*$ clearly and consistently across all evaluated scenes. The superior accuracy is largely attributed to the relative pose estimation in VMRF which is more independent of hand-crafted pose distributions as compared with the direct pose estimation in GNeRF$^*$.

\renewcommand\arraystretch{1.2}
\begin{table}[t]
\renewcommand\tabcolsep{2.0pt}
\centering 
\caption{
Ablation studies of the proposed VMRF on the scene `mic' of Synthetic-NeRF: The baseline model \textit{Base} \cite{meng2021gnerf} estimates camera poses from images directly. \textit{Base+VMS} incorporates the proposed view matching scheme (VMS) above \textit{Base}, where real-image pose is estimated from the concatenation of a rendered image with randomly initialized camera pose and a transport plan between the real and rendered images.
\textit{Base+MPC} includes the proposed pose calibration (MPC) above \textit{Base}, where MPC receives the concatenation of a real image and a rendered image with randomly initialized camera pose.
\textit{Base+VMS+MPC} is equivalent to the proposed VMRF.
All models are trained without known camera poses or hand-crafted pose distributions (the range of azimuth and elevation are randomly set to $[0^\circ, 360^\circ]$ and $[-30^\circ, 90^\circ]$).
}
\label{ablation}
\begin{tabular}{l||ccccc} 
\hline
& 
\multicolumn{5}{c}{\textbf{Evaluation Metrics}}
\\
\cline{2-6}
\multirow{-2}{*}{\textbf{Models}} 
& PSNR $\uparrow$ & SSIM $\uparrow$ & LPIPS $\downarrow$ & Rot($^{\circ}$) $\downarrow$ & Trans $\downarrow$
\\\hline

\textbf{Base} & 23.15 & 0.9270 & 0.1263 & 3.582 & 0.187  \\

\textbf{Base+VMS} & 25.03 & 0.9337 & 0.1089 & 2.735 & 0.153 \\

\textbf{Base+MPC} & 25.21 & 0.9366 & 0.1007 & 2.651 & 0.144 \\
\hline
\textbf{Base+VMS+MPC} & \textbf{27.63} & \textbf{0.9483} &  \textbf{0.0803} & \textbf{1.394} & \textbf{0.069}     \\

\hline
\end{tabular}

\end{table}


\begin{figure}[t]
\begin{center}
\includegraphics[width=0.995\linewidth]{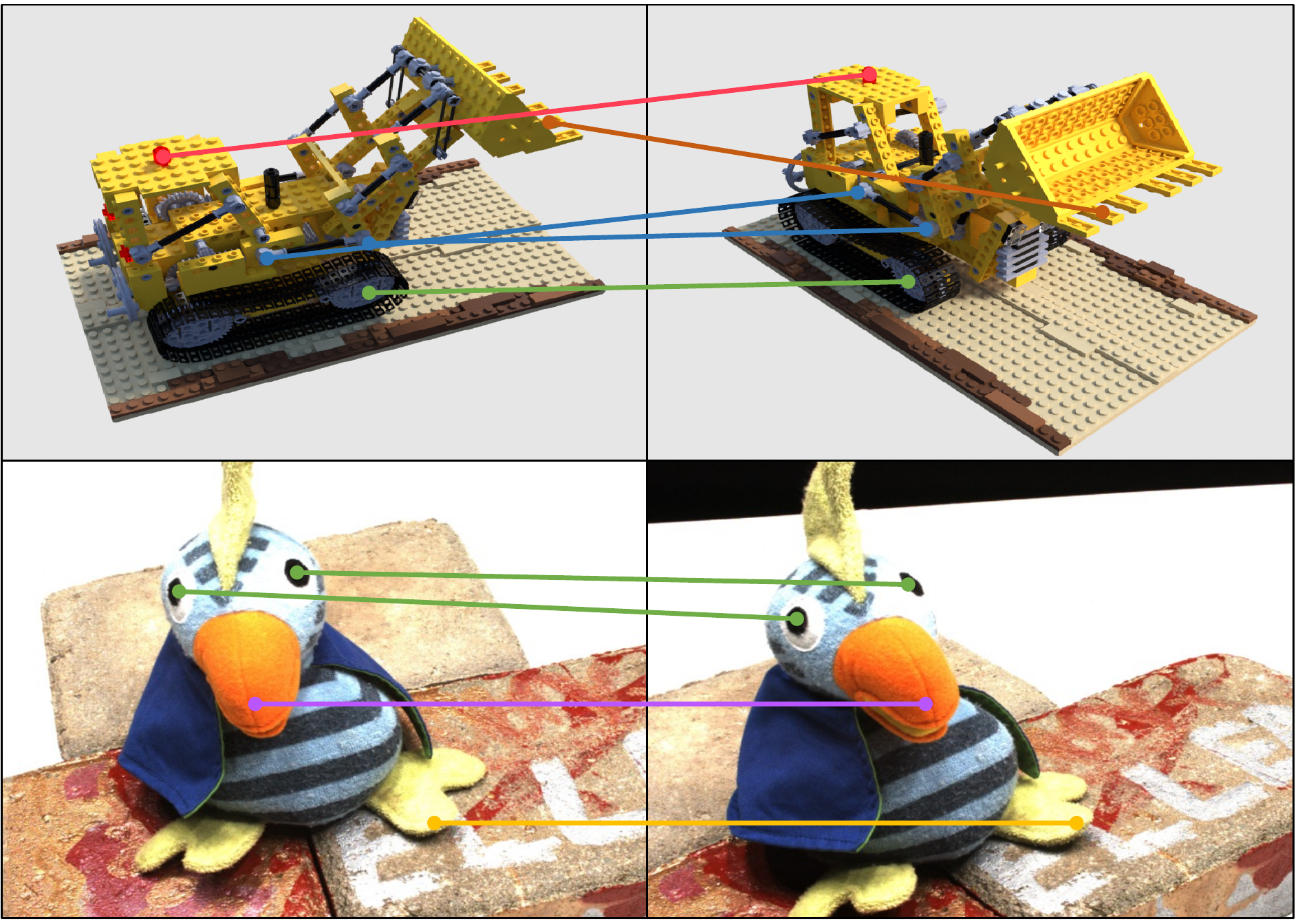}
\end{center}
\caption{
Visualization of feature matching in synthetic and real scenes: Links are drawn based on the feature matching in the derived transport plan. The upper sample is the synthetic scene `lego' from Synthetic-NeRF and the lower sample is the real scene `Scan4' from DTU. It shows that the proposed view matching with unbalanced optimal transport achieves feature-level matching effectively.
}
\label{visualization}
\end{figure}

\begin{figure}[t]
\begin{center}
\includegraphics[width=1\linewidth]{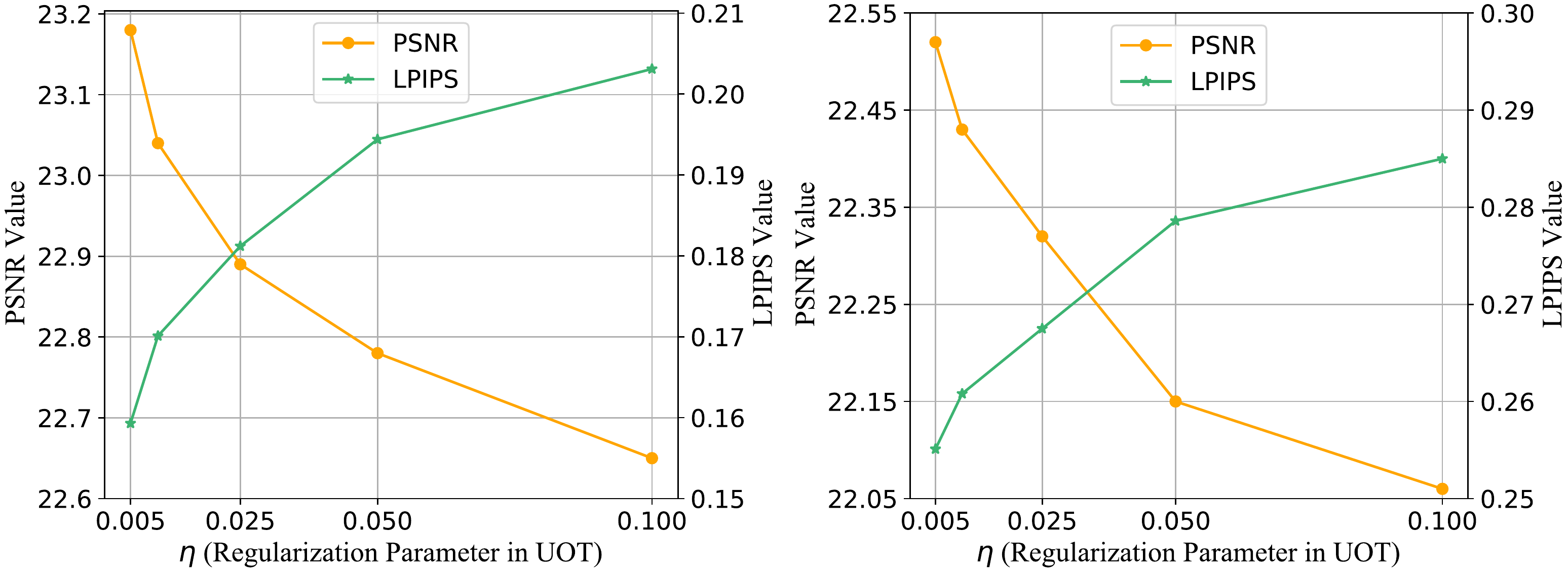}
\end{center}
\caption{
Parameter $\eta$ in VMRF: The performance of VMRF varies with $\eta$ (regularization parameter of UOT), and decreasing $\eta$ improves novel view synthesis consistently. The two graphs from left to right are computed for the scene `chair' of Synthetic-NeRF and the scene `scan63' of DTU, respectively.
}
\label{para}
\end{figure}

\subsection{Ablation Studies}

\paragraph{\textbf{Effect of relative pose estimation}} To examine the effectiveness of the proposed relative pose estimation, we perform ablation experiments to study how it affects view synthesis in PSNR, SSIM, LPIPS, Rot and Trans. As shown in Table \ref{ablation}, the model \textit{Base} works as a baseline \cite{meng2021gnerf} which directly predicts camera poses from images. The complete model \textit{Base+VMS+MPC} with both view matching scheme and matching-based pose calibration is equivalent to the proposed VMRF which performs relative pose estimation that utilizes view matching results to model relative transformation for pose calibration. Quantitative experiments show that \textit{Base+VMS+MPC} performs clearly better than the \textit{Base} in novel view synthesis, demonstrating that the proposed relative pose estimation alleviates the dependence of NeRF training on hand-crafted pose distribution effectively. The experimental results are also well aligned with qualitative experimental results in Fig.~\ref{visual_abla} where \textit{Base+VMS+MPC} with the proposed relative pose estimation generates new views with clearly less artifacts and finer details.

\paragraph{\textbf{Effect of view matching scheme}} We also conduct experiments to examine the contribution of our proposed view matching scheme. As shown in Table \ref{ablation}, we train the model \textit{Base+VMS} that includes the view matching scheme above the model \textit{Base}. Compared with \textit{Base}, \textit{Base+VMS} performs clearly better in PSNR, SSIM, LPIPS, Rot and Trans which demonstrates the effectiveness of the proposed view matching scheme. The quantitative results are well aligned with the qualitative experiments in Fig.~\ref{visual_abla} as well. Due to space limit, ablation experiments about the unbalanced optimal transport in view matching scheme are presented in the supplementary material.

\paragraph{\textbf{Effect of matching-based pose calibration}}  We further examine how the proposed matching-based pose calibration contributes to NeRF training. As Table \ref{ablation} shows,  we train the model \textit{Base+MPC} which incorporates matching-based pose calibration above the model \textit{Base}. It can be observed that \textit{Base+MPC} improves PSNR, SSIM, LPIPS, Rot and Trans consistently as compared with \textit{Base}. The better performance is largely attributed to the included relative transformation predictor that leads to robust training (with no hand-crafted pose distributions) by modeling relative pose transformation. The effectiveness of the proposed matching-based pose calibration can be observed in Fig.~\ref{visual_abla} as well where the model \textit{Base+MPC} can produce clearer visual results than the model \textit{Base}.


\subsection{Visualization}

Based on the transport plan obtained by the proposed view matching scheme, we visualize feature matching between images of different views in both synthetic and real scenes as illustrated in Fig.~\ref{visualization}. It can be observed that the features on the left image can be accurately matched to those on the right image. This shows that the proposed view matching with unbalanced optimal transport can produce accurate feature matching which provides critical guidance for the ensuing pose calibration.

\subsection{Parameters Investigation}
\label{Para_investi}
We also conduct experiments on the scene `chair' of dataset Synthetic-NeRF and the scene `scan63' of dataset DTU for studying how the regularization parameter $\eta$ affects the VMRF performance. As described in Section \ref{View Matching}, a larger $\eta$ tends to increase the smoothness of the unbalanced optimal transport. It can be observed in Fig.~\ref{para} that with the decrease of $\eta$, the performance (in PSNR $\uparrow$ and LPIPS $\downarrow$ scores) of VMRF is improved consistently. On the other hand, the stability of the model training is impaired while $\eta$ becomes too small. Therefore, we set the regularization parameter $\eta$ at $0.005$ in our implemented system.

\section{Conclusion}

This paper presents VMRF, a view matching NeRF that aims to achieve superior NeRF representations without prior knowledge in camera poses or hand-crafted pose distributions. Specifically, a view matching scheme with unbalanced optimal transport is designed to build an optimal transport plan to guide camera pose calibration. In addition, we design a matching-based pose calibration technique with a relative transformation predictor that utilizes the obtained optimal transport plan to predict relative pose transformation for camera pose calibration. With the view matching and matching-based pose calibration, VMRF achieves a novel relative pose estimation strategy that allows to reduce the dependence on hand-crafted pose distributions. Extensive quantitative and qualitative experiments over synthetic and real scenes show the superiority of proposed VMRF without prior knowledge in camera poses or hand-crafted pose distributions. Moving forward, we will focus on studying NeRF training without prior knowledge in camera poses or hand-crafted pose distributions in challenging multi-object scenes. 

\begin{acks}
This work was supported by the Ministry of Education of Singapore (Project number: MOE-T2EP20220-0003) and in part by the National Natural Science Foundation of China [grant nos. 61922064, U2033210].
\end{acks}

\bibliographystyle{ACM-Reference-Format}
\bibliography{vmrf}

\end{document}